# Application of Artificial Intelligence for Fraudulent Banking Operations Recognition

Bohdan Mytnyk[1], Oleksandr Tkachyk[1], Nataliya Shakhovska[1], Solomiia Fedushko[2,3,*] and Yuriy Syerov[2,3]

[1]
Department of Artificial Intelligence Systems, Lviv Polytechnic National University, 79000 Lviv, Ukraine
[2]
Social Communication and Information Activity Department, Lviv Polytechnic National University, 79000 Lviv, Ukraine
[3]
Department of Information Systems, Faculty of Management, Comenius University in Bratislava, 820 05 Bratislava, Slovakia

## Abstract

This study considers the task of applying artificial intelligence to recognize bank fraud. In recent years, due to the COVID-19 pandemic, bank fraud has become even more common due to the massive transition of many operations to online platforms and the creation of many charitable funds that criminals can use to deceive users. The present work focuses on machine learning algorithms as a tool well suited for analyzing and recognizing online banking transactions. The study's scientific novelty is the development of machine learning models for identifying fraudulent banking transactions and techniques for preprocessing bank data for further comparison and selection of the best results. This paper also details various methods for improving detection accuracy, i.e., handling highly imbalanced datasets, feature transformation, and feature engineering. The proposed model, which is based on an artificial neural network, effectively improves the accuracy of fraudulent transaction detection. The results of the different algorithms are visualized, and the logistic regression algorithm performs the best, with an output AUC value of approximately 0.946. The stacked generalization shows a better AUC of 0.954. The recognition of banking fraud using artificial intelligence algorithms is a topical issue in our digital society.

## 1. Introduction

The development of artificial intelligence for recognizing fraudulent banking operations has received significant attention in recent years. This is due to the growing number of fraudulent activities in the banking industry, which have resulted in significant financial losses for banks and their customers. AI-based systems have the potential to effectively identify and prevent fraudulent activities in real time, providing a significant advantage over traditional fraud detection methods.

Detecting fraudulent banking operations involves using AI and machine learning algorithms to analyze large amounts of data from multiple sources, including transaction records, customer information, and network logs. These algorithms can identify patterns and anomalies in the data that may indicate fraudulent activities, such as unauthorized access, unusual transaction patterns, and suspicious behavior.
A bank transaction involves any activity related to a bank account, which can be carried out online or offline between all parties involved. The process concludes when a written or electronic order is submitted to the bank, using internet banking systems, communication systems, or payment instruments [1]. Bank transactions fall into two categories: genuine and fraudulent transactions. The latter refers to those that violate financial circulation rules or were not authorized. Common types of banking fraud include wire fraud, identity theft, account takeover, money laundering, and accounting fraud.
As fraud becomes increasingly sophisticated, we must develop new methods to protect ourselves against it. Below are the five most common methods of preventing bank fraud: artificial intelligence, biometric data, consortium data, standardization of high technologies, and machine learning [2]. Following the outbreak of the COVID-19 pandemic and the war in Ukraine, fraudulent operations related to bank transactions have become even more common due to the significant shift toward online transactions as well as the creation of numerous charitable funds that criminals use to deceive users. Therefore, it is necessary to create reliable automated algorithms to recognize and prevent operations that threaten the finances and accounts of individuals, violate taxation or financing rules or laws, and so on. The presented study focused on machine learning algorithms as a tool well suited for analyzing and recognizing online banking transactions. This study aimed to develop machine learning models that recognize fraudulent banking transactions, especially during the COVID-19 pandemic and war in Ukraine when online transactions and charitable funds have become more prevalent.
Our project focused on using machine learning models to identify fraudulent banking transactions. We also applied preprocessing techniques to compare and select the most effective outcomes from bank data. To accomplish our goals, we took the following steps:
-Developed several machine learning models using various methodologies and strategies.
-Compared and assessed the models from the previous stage using both quantitative and visual criteria.
-Analyzed the results obtained and drew a conclusion about the research objective.
This study focused on using machine learning models to detect fraudulent banking transactions. The research aimed to develop algorithms that can accurately recognize such transactions. The methods used included preprocessing techniques and machine learning algorithms. The significance of this work lies in the potential of the proposed method to improve the detection of fraudulent banking transactions, especially during the pandemic when many transactions have shifted online and during times of war when there are many charities and events collecting money.
The task of recognizing fraudulent bank transactions using machine learning involves identifying a particular transaction at a specific moment in time as either real or fraudulent based on previous historical data about other transactions. This process uses binary logic, where the transaction is either real or fraudulent, and classification algorithms are suitable for performing the task using machine learning methods. This paper proposes applying several classification algorithms that recognize the type of transaction based on certain features, along with preprocessing techniques.
In order to effectively identify fraudulent transactions, the machine learning algorithm must have access to a comprehensive historical database of such activities. The existing collection of legitimate transactions that have yet to be flagged is encrypted to maintain the confidentiality and privacy of the financial institution's clientele. However, this encryption does not hinder the algorithm's ability to perform. Financial institutions can effectively detect and prevent fraudulent transactions by training models using this carefully selected dataset.

Implementing AI technology in detecting fraudulent banking operations poses several challenges, including the utilized algorithms' lack of transparency and interpretability. The intricacies of these algorithms can sometimes be challenging to comprehend, which can impede the identification and rectification of errors. Furthermore, using AI in fraud detection raises essential concerns regarding privacy, as personal data are subject to analysis and utilization in the decision-making process. These challenges require careful consideration to ensure AI-powered fraud detection systems' safe and accurate implementation.

The application development based on AI for fraudulent banking operations recognition is an active area of research and development, with significant potential to improve the efficiency and accuracy of fraud detection. However, addressing the challenges and concerns associated with using AI in fraud detection is essential to ensure its effectiveness and ethical use in the banking industry.

*Limitation of studies in the financial field.* Studies using artificial intelligence to detect bank fraud are valuable. However, it is important to note that this study focuses on identifying fraudulent transactions in online banking only, while other types of financial fraud may require different detection methods. Additionally, the study's reliability and generalizability may be affected by its limited sample size in the financial field. Data availability is also a crucial factor, as high-quality data re needed to train and test machine learning algorithms. The accuracy of the model can be compromised by incomplete and insufficiently diverse datasets, leading to false positives in real-world situations. Another challenge is the potential for human biases in the selection and analysis of data, which can impact the method's validity and reliability. It is also important to avoid overfitting, where the model performs well on the training dataset but poorly on the test dataset due to its complexity and limited generalizability.

# 2. Related Works

Bank fraud is stealing money or assets from a bank, financial institution, or bank depositors. Generally, bank fraud includes any act intended to defraud a financial institution. This may involve obtaining assets, loans, money, securities, or property of a financial institution due to false statements or false information. The law broadly defines bank fraud, and several aspects of this crime must be considered [3]. As mentioned earlier, there are many ways to recognize fraudulent operations. This study considered, analyzed, and compared the application of machine learning algorithms that automatically find dependencies or conclusions based on a previous historical data set. Therefore, with the help of a set of historical bank transactions, it is possible to make assumptions about a specific transaction at a given moment.

Bank fraud is a phenomenon that includes any intentional actions aimed at deceiving a financial institution or an individual. Many methods recognize such operations, but this work used machine learning methods with data preprocessing techniques.

To determine the framework of the investigations in the fraudulent banking field, we analyzed the 3111 related documents in Scopus [4] (see Figure 1).

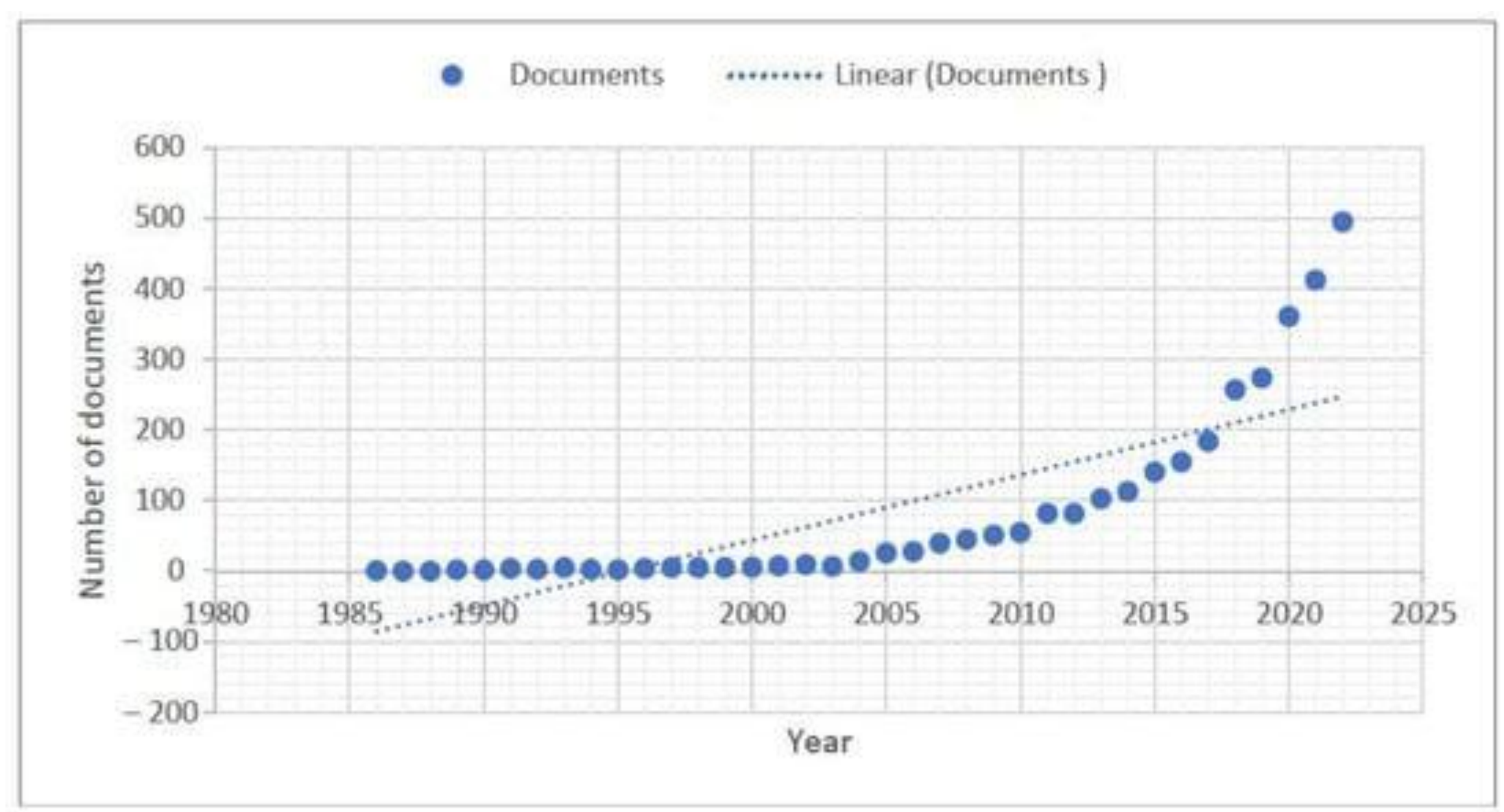

**Figure 1.** The statistics of scientific studies in fraudulent banking field in Scopus by year.
We used VOSviewer software (Centre for Science and Technology Studies, Leiden University, Leiden, The Netherlands) to create a map (Figure 2) to analyze the scientific research done in the fraudulent banking field. The map is based on the statistics of the average number of publications in Scopus each year and the analysis of 348 Scopus documents. The map has 827 items, 3 clusters, 94,199 links, and a total link strength of 263,429.

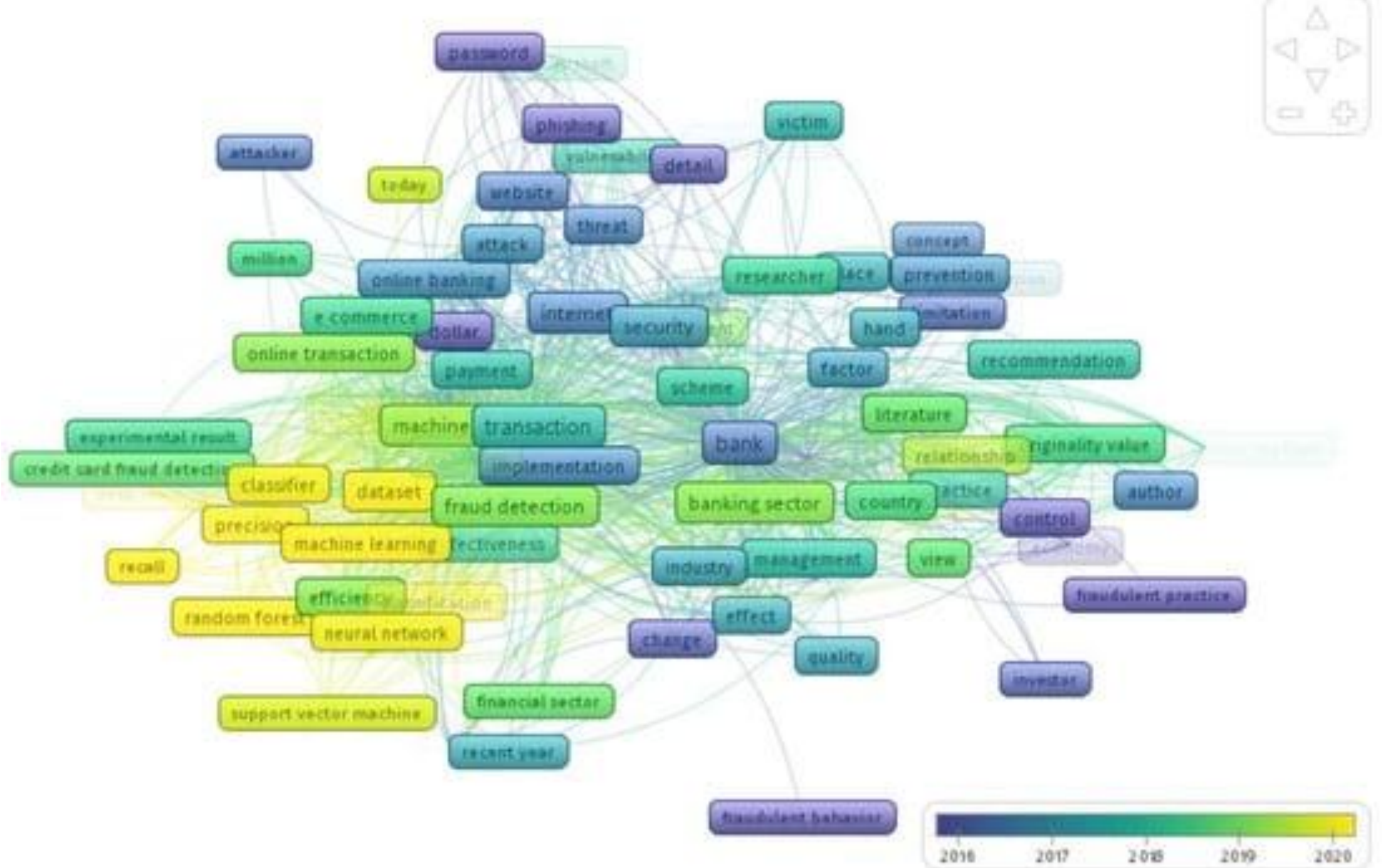

**Figure 2.** Map based on statistics of average number of publications in Scopus each year.
The map (Figure 2) shows that fraudulent banking practices have become a significant concern for financial institutions and regulatory bodies worldwide. These practices involve deliberate attempts by individuals or organizations to deceive banks, financial institutions, or their customers for financial gain. Fraudulent activities in banking can take various forms, such as phishing scams, identity theft, money laundering, and account takeover fraud.

Fraudulent banking practices such as phishing scams and identity theft are unfortunately common. Phishing scams involve deceiving customers through fraudulent emails, text messages, or phone calls to obtain their personal and financial information. Perpetrators often use sophisticated techniques such as creating fake websites that appear legitimate to trick customers. Identity theft occurs when someone's personal information is stolen and used for fraudulent transactions or obtaining credit. The effects of identity theft can be long-lasting, causing significant financial and emotional damage that may take months or even years from which to recover.

Money laundering is another fraudulent banking practice involving concealing the origin and ownership of illegally obtained funds. This practice is often associated with organized crime and can have profound implications for financial institutions that are found to have facilitated the laundering of illicit funds.

Account takeover fraud is a fraudulent banking practice when a criminal gains unauthorized access to a customer's bank account. This can be achieved by stealing the customer's login credentials or exploiting vulnerabilities in the bank's security systems. Once the criminal gains access to the account, they can make fraudulent transactions or steal funds.

Financial institutions and regulatory bodies have taken steps to prevent fraudulent banking practices. These include advanced fraud detection technologies, enhanced customer verification procedures, and increased regulatory oversight. As fraudulent activities are always changing, it is important for these bodies to remain alert and adaptable to continue combating fraudulent banking practices.

Fraudulent banking practices pose a serious threat to the stability and integrity of the financial system. Financial institutions and regulatory bodies must work together to develop effective strategies and measures to prevent and detect fraudulent activities. Only through a coordinated and proactive approach can the financial system be protected from the harmful effects of fraudulent banking practices.

The fraudulent banking approach and threats are shown in Table 1.

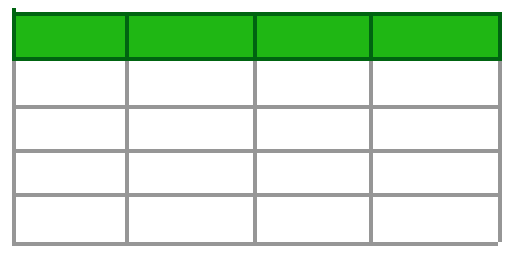

**Table 1.** Fraudulent banking approaches and threats.

Some approaches may involve a combination of tactics. Banking clients need to stay vigilant and protect themselves from fraud, such as regularly monitoring their accounts and avoiding clicking on suspicious links or downloading unknown software.

A study [13] proposed a mechanism for detecting credit card fraud based on machine learning, which uses a genetic algorithm to select features. After selecting the best features, the following classifiers are used: random forest, artificial neural network, decision tree, logistic regression, and naïve Bayesian network. The efficacy of this approach, which relies on data from European cardholders, surpasses that of existing methods, according to this study. Furthermore, it is notable for its similarity to another method that emphasizes the significance of preprocessing algorithms. It is worth mentioning that the method exclusively employs the genetic algorithm as its preprocessing algorithm.

Some researchers [14] aimed to develop an unbiased, reliable, and easy-to-use methodology for automatically assessing card fraud risk. Thus, a new methodology was proposed that uses algorithms that quantify information about variables and their relationships. The authors also used a state-of-the-art recurrent filters set to minimize the training data bias, i.e., a filter for repeated features and a filter for the most informative features. Subsequently, the outcomes were categorized by applying machine learning techniques, including linear discriminant analysis, support vector machine, gradient boosting, and linear regression. The identified models were applied to synthetic and real databases. As a result, 76% accuracy was obtained, which is a relatively good number for such a task but not the best. Nevertheless, a considerable advantage of this method is its proposed new methodologies for data preprocessing.
A paper [15] proposes the use of a multilayer perceptron to improve the accuracy of the credit card fraud detection process. This study measured the method's effectiveness based on accuracy, specificity, precision, sensitivity, root mean squared error, area under the curve, and F-measure. The experimental outcomes demonstrated that the suggested model, which employs an artificial neural network, showed significantly enhanced precision in identifying fraudulent transactions. The advantage of this study is the large number of different evaluation measures, which allow the readers to more broadly consider the effectiveness of the shown solution.
Another methodology [16] centers on differentiating fraudulent credit cards from fraudulent transactions. The concept of the suggested model is user segmentation, which partitions users into new and old categories; subsequently, CatBoost and a deep neural network are applied to separate categories correspondingly. This study also elaborates on multiple techniques to enhance the precision of detection, namely managing imbalanced datasets, transforming features, and engineering features. The experimental results showed that the AUC was 0.97 for CatBoost and 0.84 for the deep neural network. The advantage of the method is a nonstandard approach to the problem, which focuses on users rather than transactions and, as a result, achieves good accuracy.
An article [17] proposes a method to detect credit card fraud using a combination of an ensemble neural network classifier and a hybrid data resampling technique. The ensemble classifier uses a long short-term memory (LSTM) neural network and the AdaBoost algorithm. The study evaluated the effectiveness of this approach by comparing it with other machine learning algorithms. The results showed that the classifiers performed better when trained with repeatedly sampled data. LSTM ensemble performed better than other algorithms, achieving a sensitivity and specificity of 0.996 and 0.998, respectively. The method's high accuracy is a notable advantage. In a study [18], several machine learning algorithms, such as logistic regression, support vector machine, neural networks, and random forest, were employed to train a machine learning model based on a given dataset. The authors conducted a comparative study of these algorithms' accuracy and other performance metrics. The study concluded that the artificial neural network performed the best, obtaining an F1 score of 0.91. This work is also noteworthy for its relevance to the current problem.
Machine learning techniques for automated credit card fraud detection generally do not account for fraud sequences or behavioral changes that may result in false alarms. Thus, authors [19] proposed a detection system for credit card fraud that uses LSTM networks, utilizing a learning system to incorporate transaction sequences. The suggested method aims to capture historical credit card purchasing behavior to enhance fraud detection accuracy for incoming transactions. The experimental outcomes showed that the suggested model yields robust results, and its accuracy is relatively high. A significant advantage of this method is its focus on predicting fraudulent transactions and minimizing false positives, a factor rarely addressed in the existing literature on this subject.

In reference [20], a study was conducted to detect credit card fraud using machine and deep learning methods in the healthcare sector. These algorithms include naïve Bayes, sequential convolutional neural network, KNN, logistic regression, and random forest. The accuracy rates of each algorithm were as follows: naive Bayes, 96.1%; logistic regression, 94.8%; KNN, 95.89%; random forest, 97.58%; and convolutional neural network, 92.3%. However, the overall comparative analysis revealed that the KNN algorithm outperformed the other approaches, which was somewhat unexpected given the convolutional neural network and random forest in the algorithm list.

A paper [21] proposes a framework to handle metrics for strings. The framework was applied to generalize the edit distance metric. The scientists investigated the computational properties and solution algorithms for the multiparameterized edit distance, performed experiments for its evaluation, and discussed the possible applications of the multiparameterized edit distance and other generalized metrics in various scenarios.

A study [22] monitored air quality using IoT devices. The study proposed a mixed edge-based and cloud-based framework for PM2.5 value prediction and evaluated the framework's quality using a real-world dataset. The proposed preprocessing technique showed an average upgrading of 40.18% in prediction accuracy on the mean absolute percentage error (MAPE).

The literature shows that detecting fraudulent financial transactions using artificial intelligence is an important topic. A study [23] highlights the importance of secure collaborative information systems in organizations and how AI, deep learning, and blockchain technologies are being used to secure these systems. The paper presents a model for detecting fraud and authenticating users. The logistic regression technique was used to create a regression model for participant authentication.

A paper [24] proposes a quality, experience-based web platform to improve bank client satisfaction and offer quality service. The platform allows customers to enter complaints and information to analyze an employee's performance and behavior with customers.

A paper [25] discusses the increase in credit card fraud transactions and the need for banks and credit card businesses to classify fraudulent transactions to protect customers. The study used machine learning approaches to detect credit card fraud. The Random Oversampling technique yielded the best results with a precision and accuracy score of 0.99. The researchers proposed the implementation of data sampling methods as means of balancing data for optimization of the performance of the model in effectively classifying fraudulent activities.

In an article [26], the use of data science and machine learning was explored to detect credit card fraud. The project focused on creating an AI system that can detect fraud even in imbalanced datasets. To achieve this, the team emphasized the importance of feature engineering and dataset modification. They also acknowledged the challenge of adapting the system to real-time situations, given the high volume of credit card transactions. The article provides details on the evaluation metrics and machine learning techniques used to differentiate between each analysis.

A study [27] investigated the factors affecting the bank's intention to adopt internet banking using the technology model. The study identified the factors that exhibited statistical significance in predicting bank clients' probability of adopting banking systems.

Investigators [28] discussed the increase in cyberattacks on financial institutions and the need for better modeling techniques to mitigate them. Financial sector cyberattack modeling was carried out using the Bayesian attack network modeling technique, which utilizes phishing emails and exposures to obtain conditional probabilities for the modeling process. The method employs the generation of probability density curves for countless attack structures and exploits the degree of exploitability to mitigate attacks.

Another study [29] used a questionnaire to collect data from bank employees in different positions. According to the findings, the management level of knowledge implementation was moderate regarding knowledge creation and acquisition, with a score of 68%. On the other hand, it was high in the storage of knowledge (74%), application (77%), and sharing (76%).

Many authors have attempted to compare different machine learning algorithms for this purpose but have taken different approaches to the problem, demonstrating the importance of this topic and its practical application in the banking industry.

# 3. Materials and Methods

To meet the goals outlined in this study, we employed classification algorithms. These algorithms use features to determine the class of a given object. Machine learning relies on labeled training data to categorize new observations. To make accurate predictions for future observations, these algorithms must first analyze a dataset of examples with features and corresponding classes. They are considered supervised learning techniques because they map input variables (x) to discrete output functions (y) that represent categories rather than numerical values.

The output of classification algorithms is not continuous but discrete. The classification algorithm learns from labeled input data, where input data have a corresponding output. The goal of classification algorithms is to limit the category of a given dataset, and they are widely used to predict the output for categorical data [30].

From the information provided in the previous paragraph, it is clear that training any classification model requires a data set that shows the relationship between certain feature sets of an object and its class or category. This is why, to fit our classification model to the given task, we chose to use the dataset named Credit Card Fraud Detection obtained from the Kaggle platform [31]. The platform provides the ROC graph curve and AUC metric for each algorithm and technique. A similar set of metrics is provided for other implementations as well. It is also easy to run the program using the Kaggle platform (which was used to create this software solution), which, by default, supports running Jupyter notebooks.

To start, the “Run all” was pressed button to continue the sequential execution of all commands. An alternative solution is using any software that supports Jupyter notebooks. The concrete program implementation as saved as a notebook on the Kaggle platform and started by loading a dataset using the Pandas library.

The research workflow was organized as given in Figure 3. Figure 3 represents the stages of a high-level algorithm for a machine learning program solution, which includes dataset selection and loading, feature standardization, random undersampling, model fitting, model testing, and outputting the best model”.

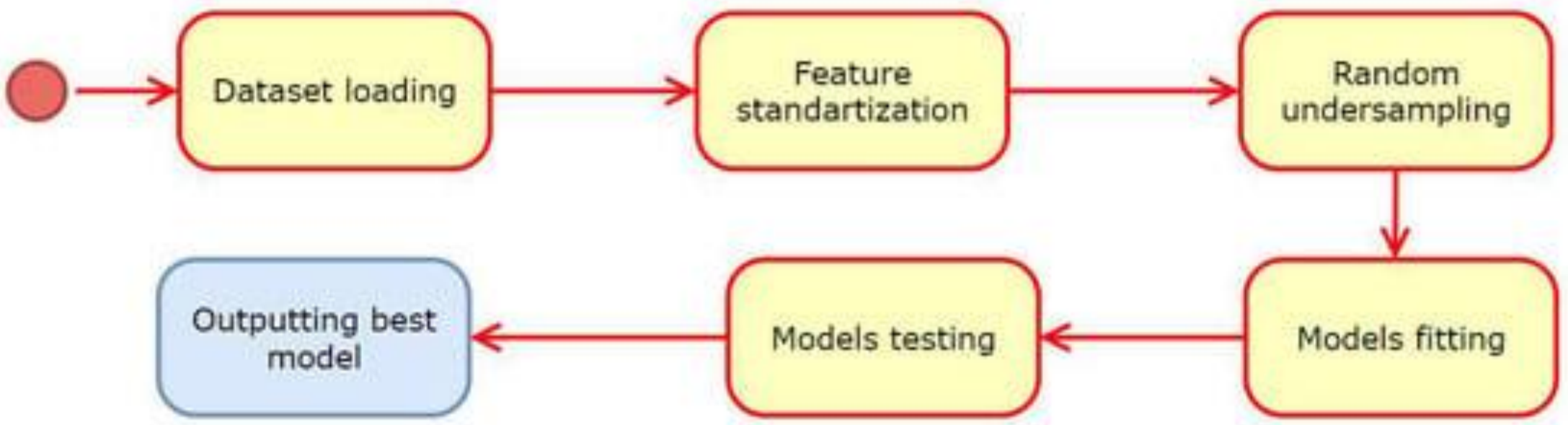

**Figure 3.** Scheme of the program solution to the problem.

- Selecting the dataset using Kaggle datasets.
- Loading the dataset into the program (using a library such as pandas or NumPy to load the dataset into the program).
- Splitting the dataset into training and testing sets (using a library such as scikit-learn).
- Standardizing the features in the training and testing sets using a standard scaler.
- Imbalanced learning to randomly undersample the majority class in the training set to balance the class distribution.
- Selecting a set of candidate models to evaluate.
- For each model, fitting the model to the training data using hyperparameter tuning (using a library such as scikit-learn), which involves using cross-validation to find the best hyperparameters for each model.
- Using a library such as scikit-learn to evaluate each model on the testing data using an appropriate evaluation metric.
- Recording each model's evaluation metric and hyperparameters for comparison.
- Selecting the model with the best evaluation metric on the testing data.
- Outputting the best model and its hyperparameters and evaluation metric for further use in production or research.

This algorithm provides a framework for building a machine learning program solution that includes dataset loading, feature standardization, random undersampling, model fitting, model testing, and outputting the best model. The presented method uses several mathematical elements:

- Machine learning algorithms,
- Evaluation metrics,
- Data preprocessing technique.

### *3.1. Machine Learning Algorithms*

The machine learning algorithms chosen to be used here were (1) random forest, (2) k-nearest neighbors, (3) logistic regression, (4) stochastic gradient descent classifier, (5) decision tree, (6) naïve Bayes, and (7) support vector machine.

*Decision tree.* In machine learning, a decision tree is a tree structure that shares similarities with a flowchart. Each internal node of the tree represents an attribute check, while each branch corresponds to the outcome of the check. Finally, every end node, or final node, contains a class label. The initial set is divided into subsets to train the decision tree based on checking the attribute's value. The described process is recursive partitioning, repeated recursively for each derived subset. The recursion split ends in case of partitioning are no longer beneficial for predictions. Classification based on decision tree methods does not require knowledge of the domain or parameter tuning, making it an excellent choice for exploring knowledge. Furthermore, decision trees handle large amounts of data and typically provide high accuracy. Decision tree induction is a standard inductive approach for learning classification data. To classify instances, decision trees [32] sort them through the tree, starting at the root and ending at a leaf node that classifies the instance.

*Random Forest.* Specialists have used random forest methods for classification and regression tasks. This machine-learning-based algorithm is based on a flexible and user-friendly algorithm consisting of decision trees. The strength of the forest increases with the number of trees. The algorithm creates decision trees using randomly selected data samples, obtains predictions from each tree, and selects the best solution by voting.

Additionally, it indicates feature importance. The algorithm works in the following steps: (1) selecting random samples, (2) building a decision tree, (3) voting, and (4) selecting the prediction result as the final prediction [33].

*Logistic regression* is a commonly used statistical model for classification and predictive analytics, estimating the probability of an event based on a given set of independent variables [34]. Logistic regression transforms the odds using a logit transformation, the logarithm of the odds, or the natural logarithm of the odds.

*Support vector machine (SVM)* is a supervised learning algorithm for classification tasks and regression assignments [35]. SVM creates a decision boundary, or hyperplane, that divides n-dimensional space into classes. The hyperplane is created by selecting extreme points, or support vectors, that help define the boundary.

*K-nearest neighbors (KNN)* is a supervised learning nonparametric classifier. This classifier utilizes proximity to classify or predict the grouping of an individual data point. KNN is commonly used as a classification algorithm and assigns a class label based on the majority vote of nearby data points. For classification tasks with multiple classes, the class label is assigned with more than 25% of the vote rather than a strict majority of over 50% [36].

*The stochastic gradient descent (SGD)* classifier is an approach that is straightforward and remarkably efficient in adapting linear classifiers and regressors to convex loss functions, including logistic regression and support vector machine. However, SGD has recently received substantial attention in the realm of large-scale learning. This is due to the success of SGD in tackling vast and sparse machine-learning challenges commonly encountered in natural language processing and text classification. Due to the sparsity of the data, the classifiers developed using SGD have rapidly expanded to deal with problems with over $10^5$ training examples and over $10^5$ features [37].

*The naïve Bayes classifier* is a probabilistic machine learning model for classification tasks. The classifier's foundation is based on Bayes' theorem. The naïve Bayes classifier assumes that all predictors or features are independent, meaning that the presence of one feature does not affect the other [38]. The Bayes formula is represented as the following equation:

$$P(A|B)=P(B|A)\times P(A)P(B), \quad (1)$$

where $P(A)$ is the event $A$ probability, $P(B)$ is the event $B$ probability, and $P(B|A)$ is the probability of event $B$ occurring when event $A$ occurs.

In probability theory, the probability of an event $A$ occurring is denoted by $P(A)$, while the probability of event $B$ occurring is denoted by $P(B)$. Furthermore, the conditional probability of event $B$ happening certainly that event $A$ has happened is represented as $P(B|A)$.

### 3.2. Stacked Generalization of Machine Learning Models

Stacked generalization is a widely used method that integrates multiple low-level models to improve the overall predictive accuracy of a high-level model. This technique is based on estimating the biases of the high-level model concerning a given learning dataset. The estimation process involves generalizing the biases in a second space, using the original models' predictions as inputs and the correct answers as outputs. Stacked generalization [39] is considered an enhanced version of cross-validation, aggregating individual models into a higher-level model. Recently, a new method for grouping machine learning models based on random forest as a meta-algorithm has been developed, and its mathematical formulation is presented below.

Randomly generate the following from the original dataset $K$ cross-sectional data sets:

$$\{a_{11},\ldots,a_{1B}\},\{a_{21},\ldots,a_{2B}\},\ldots,\{a_{K1},\ldots,a_{KB}\}, \quad (2)$$

where $K$ is the number of subsets, $B$ is the size of the subset, and $a_{lb}$ is the observation of the $l$th sample.

The task is to train $K$-independent weak classifiers

$$f_1(.),\ldots,f_k(.)$$

Furthermore, combine the learning results using metamodel m:

$$res=m(f_1(.)\times f_2(.)\times\ldots\times f_k(.))$$

where $f_i(.)\times f_j(.)$ is the result of the pairwise multiplication of weak classifiers.

The transformed features are combined with the training dataset in the metamodel to improve model generalizability and prevent the correlation of weak classifiers' results. However, the stacking model has a significant drawback: the meta-attributes for the training and testing sets differ. The meta-attribute in the training set is not the response of a specific classifier; it comprises responses from various classifiers with different types of dependence. On the other hand, the meta-attribute in the testing set is the answer to a completely different classifier configured for complete learning. The meta-attribute may have few unique values in classical stacking, but many do not overlap between the training and testing sets.

### 3.3. Metrics for Model Assessment

We employed the receiver operating characteristic (ROC) curve and the area under the curve (AUC) to evaluate the efficiency of the models suggested in this article. The ROC curve is a graphical representation of a classification model's accuracy for all classification thresholds. This curve is plotted using the true positive rate ($TPR$) and false positive rate ($FPR$). $TPR$ is the recall measure defined as the ratio of true positives to the sum of true positives and false negatives.

$$TPR=\frac{TP}{TP+FN}, \quad (3)$$

where $TP$ is true positive model labels; $FN$ is false negative model labels.

In model evaluation, $TP$ refers to the number of positive instances correctly identified by the model, while $FN$ represents the number of positive instances incorrectly classified as negative by the model.

The $FPR$ is a measure employed in assessing model performance, which quantifies the proportion of negative instances incorrectly classified as positive by the model concerning the total number of negative instances that include both the truly negative samples and the ones misclassified as positive. The $FPR$ is defined as follows:

$$FPR=\frac{FP}{FP+TN}, \quad (4)$$

where $TP$ is true positive model labels; $FN$ is false negative model labels.

In the context of model evaluation, $TP$, which represents the true positive model labels, is the count of positive instances correctly identified by the model. $FN$, which stands for false negative model labels, indicates numerous positive instances incorrectly classified as negative by the model.

The ROC curve plots the relationship between the $TPR$ and the $FPR$ at varying classification thresholds. By reducing the classification threshold, more items are classified as positive, increasing numerous true and false positives. A standard ROC curve is commonly used in evaluating model performance [39].

The area under the curve (AUC) is a widely used metric in evaluating binary classification models. It offers a comprehensive measure of performance for all possible classification thresholds. The AUC is interpreted as the model's probability of assigning a higher score to a random positive instance than a random negative instance. A perfect model has an AUC of 1.0, and the model with random predictions has an AUC of 0.0. The AUC is advantageous as it is scale-independent and invariant to the classification threshold, enabling the comparison of different models across different datasets. However, the scale invariance and threshold invariance of the AUC may not always be desirable in certain use cases. For instance, the AUC may not be applicable in situations where well-calibrated probabilistic input data are needed. Similarly, when significant differences exist in the costs of false positives and false negatives, the AUC may not be the most appropriate metric. For example, minimizing false positives in spam detection may be more important than minimizing false negatives [40], which the AUC does not consider.

## *3.4. Data Preprocessing Techniques*

In this study, the preprocessing stage involved standardizing certain features not encrypted in the dataset, specifically, the time and amount variables of submitted transactions. In addition, undersampling was employed as the dataset was not imbalanced. Standardization [41] is a technique for scaling variables where values are centered around the mean and have a standard deviation of one. Thus, the attribute's mean value is transformed to zero, and the distribution is normalized with a standard deviation of one. The standardization equation is expressed as follows:

$$X'=\frac{X-\mu}{\sigma}, \quad (5)$$

where $\mu$ is the mathematical expectation, and $\sigma$ is the standard deviation.

In order to achieve the tasks set in this study, classification algorithms needed to be employed. A technique called random undersampling was utilized, which involves the random selection and removal of examples from the majority class in the training dataset. This results in a decrease in the number of many examples in the common class in the transformed training dataset. This procedure is repeated until the desired class distribution is obtained, such as an equal number of examples for each class [42].

Classification is a supervised learning method used to determine the class or label of a given observation based on a dataset of previous observations. The dataset used in this paper was obtained from the open platform Kaggle and contained data on transactions of European cardholders labeled as fraudulent or real. Most of the features of this dataset are encrypted using the principal component method to ensure user privacy, with the only open features being the time of the transaction and the amount involved.

Preprocessing was applied to the dataset in the form of standardization of features that were not encrypted as well as the balancing of classes by random sampling.

The program's output is a graph of the ROC curve for each selected algorithm and the AUC metric, which allows quantitative, not just visual, evaluation of the algorithm's effectiveness. The demonstrated flowchart was used for the programmatic realization of the goal.

Table 2 represents the results of the analysis of the pros and cons of each algorithm. The effectiveness of algorithms varies depending on the specific problem and dataset being used. The performance of each algorithm improved with the proper tuning of the hyperparameters and feature engineering.

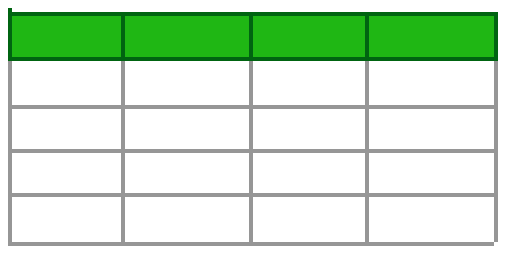

**Table 2.** The initial state of the algorithms.

Among the selected machine learning algorithms that were used for training and prediction were (1) random forest, (2) k-nearest neighbors, (3) logistic regression, (4) linear discriminant analysis, (5) decision tree, (6) naïve Bayes, and (7) support vector machine.

# 4. Results

## *4.1. Dataset Selection*

Three datasets were taken into account:

- Credit Card Fraud Detection with 150.83 Mb (https://www.kaggle.com/datasets/mlg-ulb/creditcardfraud (accessed on 15 November 2022)). This dataset presents 284,807 transactions that occurred in two days, and only 492 of them have frauds. It means that this dataset is highly unbalanced.
- Credit Card Fraud with 76.28 Mb (https://www.kaggle.com/datasets/dhanushnarayananr/credit-card-fraud (accessed on 15 November 2022)). This dataset is simulated, which is why the accuracy in one of the solutions (URL: https://www.kaggle.com/datasets/dhanushnarayananr/credit-card-fraud/discussion/335338 (accessed on 15 November 2022)) is one.
- Fraud Detection—Credit Card with 102.92 Mb (https://www.kaggle.com/datasets/yashpaloswal/fraud-detection-credit-card (accessed on

15 November 2022)). This dataset is obtained from the first dataset by removing missing values. This is why the first dataset was used in the study.

In addition, more than 4050 notebooks were developed based on the mentioned dataset. This allowed us to compare our results with those of existing methods.

### *4.2. Experiment Design*

The technical implementation of the task outlined in this paper was conducted using the high-level programming language Python [43]. The simplicity, consistency, flexibility, powerful AI and ML libraries and frameworks, platform independence, and the large community of Python are widely recognized as the ideal solution for machine learning and AI-driven projects. The development environment utilized for this project was notebooks [44] from the Kaggle platform, where the dataset was obtained. Notebooks are composed of a sequence of cells that can be formatted in either Markdown for text or a programming language of the user's choice for code [44].

In the technical implementation of this task, the following libraries were utilized:

- Scikit Learn is an open-source machine learning library that supports both supervised and unsupervised learning. Scikit Learn also provides various tools to adapt models, preprocess data, select models, and assess models, among many other services [45].
- Pandas is an open-source library primarily intended to conveniently and efficiently handle labeled or relational data. It offers several data structures and functionalities that enable numerical data and time series processing. This library was developed on the foundation of the NumPy library and is known for its fast performance and high productivity for users [46].
- Matplotlib is an open-source library that enables data visualization and plotting for Python, and it supports its numerical extension NumPy. This library provides a feasible alternative to MATLAB and is compatible with different operating systems. Creators use the matplotlib API to embed graphs in GUI applications [47]. Because the program is implemented and stored in Kaggle, an online platform, the code of the software solution can be concurrently run in the browser of the application's end user. This is why the need for personal computing power is not critical for running programs available to any modern computer with a sufficiently good Internet connection.

This dataset is readily available on the same platform, facilitating rapid processing. Once the data are loaded into memory as a frame data structure, they are sequentially processed using the earlier preprocessing techniques: standardization and random undersampling. Standardizing the features in the training and testing sets is performed by a standard scaler, and random undersampling is used to balance the class distribution. The selected models use a set of hyperparameters.

The dataset was split into training, validation, and testing sets (70%, 15%, and 15%, respectively) to measure the generalization performance. Decision tree, logistic regression, SVC, k-nearest neighbors stochastic gradient descent, naïve Bayes, and random forest algorithms were used for classification task solving. Next, grid search for hyperparameters' tuning was used. After the initialization of the models and data processing, their successive training and evaluation using the AUC metric began and the ROC curve was visualized for each of the algorithms. After the models performed the given task, the program displayed the algorithm with the best result based on the AUC metric.

Figure 4 shows the ROC curves for these algorithms and their metrics:

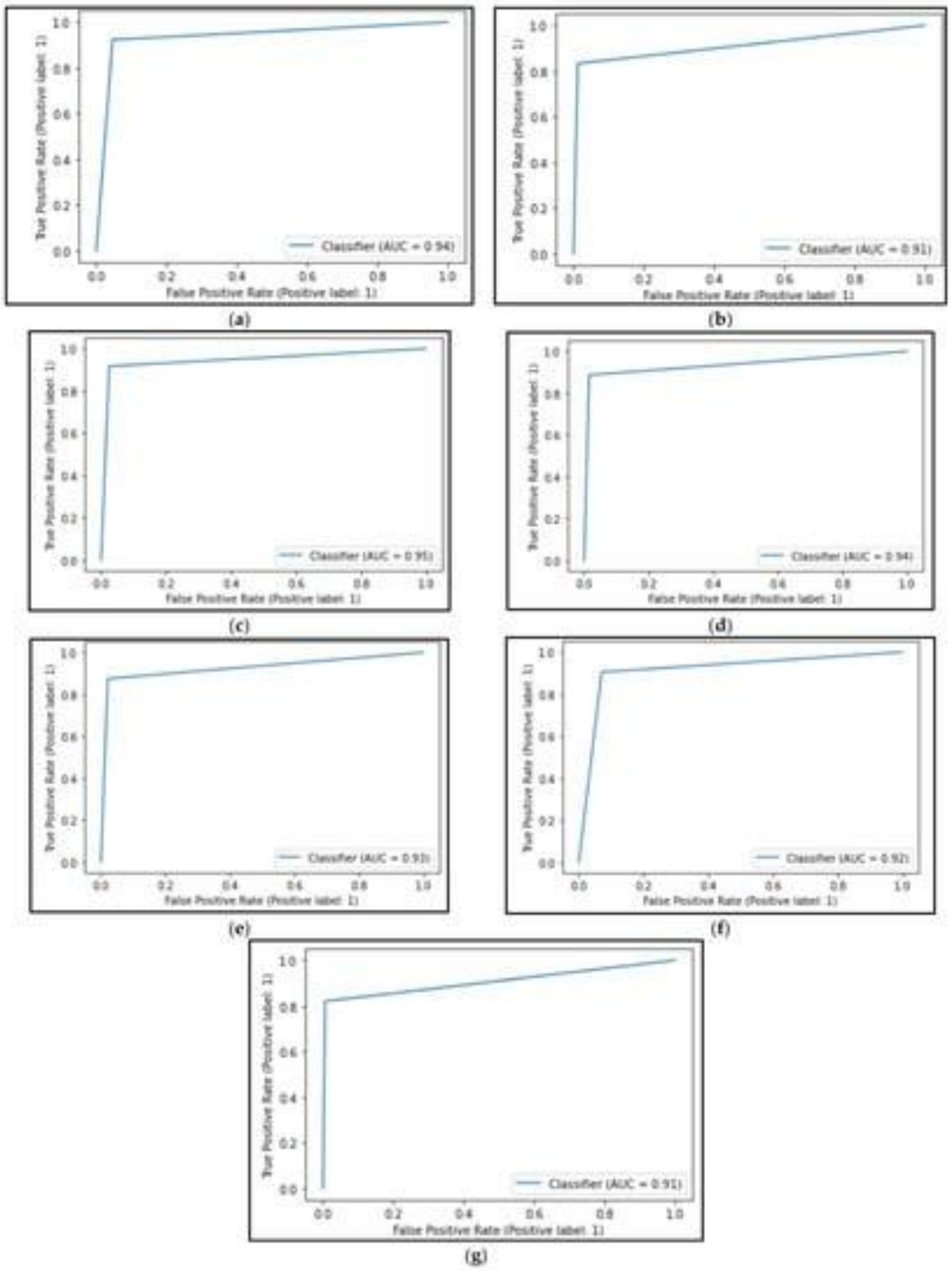

**Figure 4.** Plots of ROC curves of the following algorithms: (**a**) decision tree algorithm; (**b**) random forest algorithm; (**c**) logistic regression algorithm; (**d**) SVC algorithm; (**e**) k-nearest neighbors algorithm; (**f**) SGD algorithm; (**g**) naïve Bayes algorithm.

- The decision tree algorithm produced an AUC metric of 0.938.
- The logistic regression algorithm obtained an AUC value of 0.946.
- The SVC algorithm showed an AUC of 0.936.
- The k-nearest neighbors algorithm showed an AUC value of 0.927.
- The algorithm based on stochastic gradient descent had an AUC value of 0.917.
- The naïve Bayes algorithm showed an AUC value of 0.908.
- The random forest algorithm had an AUC value of 0.911.

Figure 5 shows the overall output of the program, which shows the best algorithm according to the AUC metric, namely, logistic regression, which obtained an AUC value of approximately 0.946.

```
Best classifier is LogisticRegression with AUC = 0.946025641025641
```

**Figure 5.** The initial state of the program.

However, stacked generalization produced better results than logistic regression. The summary table of the results is presented in Figure 6. The AUC and F1 score were used for model evaluation. The results of stacked generalization was compared with state-of-the-art results, URL: https://www.kaggle.com/datasets/mlg-ulb/creditcardfraud/code (accessed on 15 November 2022) [31] and https://www.kaggle.com/code/janiobachmann/credit-fraud-dealing-with-imbalanced-datasets (accessed on 15 November 2022) [48]. The highest F1 was calculated for the ensemble model at 0.96.

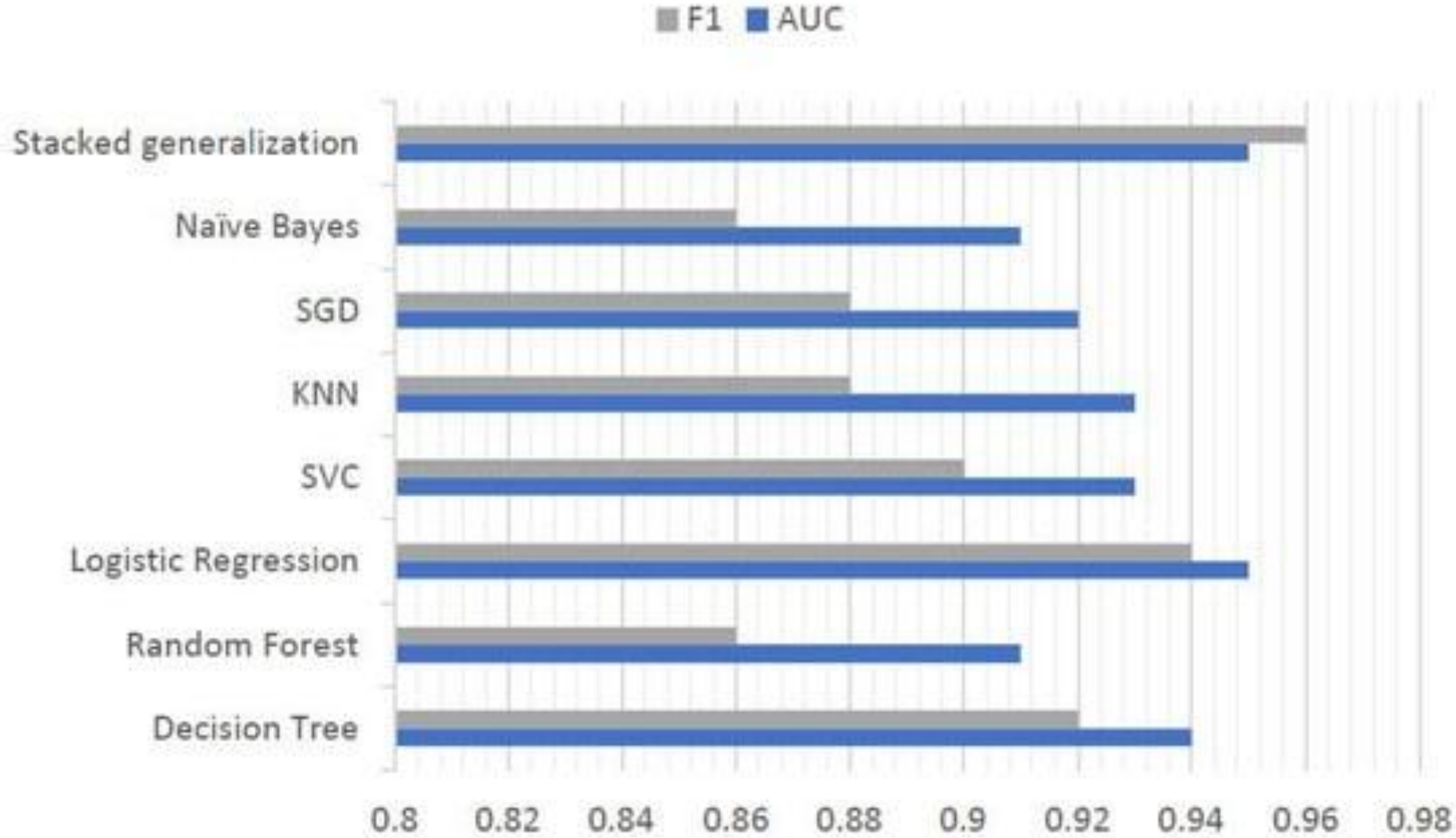

**Figure 6.** Summary of results.

The easiest solution to run the program is to use the Kaggle platform (which was used to create this software solution), which, by default, supports running Jupyter on laptops. To start it, click the "Run all" button, which continues to execute all commands one by one. An alternative solution is to use any software that supports launching Jupyter laptops. This study used the root programming language Python to create a software solution because of its unique adaptation to artificial intelligence tasks and auxiliary libraries for this language: Pandas, Sklearn, and Matplotlib.

The program was created and published on the Kaggle platform, so it does not depend on the end user's computing power and only requires a stable Internet connection and a web browser. From the test runs of the presented algorithms, we concluded that they all more or less equally cope with recognizing bank fraud. This is also evident from the ROC curve plots, which do not show much visual difference. Nevertheless, based on the numerical metric AUC, we can see that the logistic regression algorithm performed best from weak classifiers, with an output AUC value of approximately 0.946.

## 5. Conclusions

This paper emphasized the importance of utilizing artificial intelligence to identify fraudulent banking transactions. We proposed various classification algorithms that can determine the type of transaction based on specific features. The proposed model, which is based on an artificial neural network, significantly increases the accuracy of detecting fraudulent transactions. Additionally, the paper provided multiple methods for enhancing detection accuracy, such as managing imbalanced datasets, feature transformation, and feature engineering.
This paper presented the recognition of banking fraud using artificial intelligence algorithms. As a result of training and testing, each selected algorithm showed outstanding (AUC values were not lower than 0.9 in any of the cases) and equal results. This could also be seen from the graphs of the ROC curves, which do not show a significant visual difference. From the test runs of the presented algorithms, it was concluded that all of them more or less equally cope with recognizing fraudulent bank transactions. Nevertheless, based on the numerical AUC metric, the logistic regression algorithm performs best, obtaining an AUC value of approximately 0.946. The stacked generalization with deformed results of the weak classifier was proposed in the paper with an AUC of 0.008, being better than the best weak classifier. On the other side, stacking reduces bias and variance, but it is exceptionally efficient at preventing overfitting and variance. The improvement provided by the linear stacking model over the best individual model was relatively small. There was often no improvement, especially in cases where the individual base model was already sophisticated, e.g., gradient-boosted trees. The particular model's output deformation was used in the stacked generalization.
This study is important because we applied artificial intelligence to identify fraudulent banking transactions. This is particularly relevant during the pandemic, as more transactions are performed online, and during times of war, when there are many charities and events collecting money.

## Author Contributions

Conceptualization, B.M., O.T., N.S., S.F. and Y.S.; methodology, B.M., O.T., N.S., S.F. and Y.S.; software, B.M., O.T., N.S., S.F. and Y.S.; validation, B.M., O.T., N.S., S.F. and Y.S.; formal analysis, B.M., O.T., N.S., S.F. and Y.S.; investigation, B.M., O.T., N.S., S.F. and Y.S.; resources, B.M., O.T., N.S., S.F. and Y.S.; data curation, B.M., O.T., N.S., S.F. and Y.S.; writing—original draft preparation, B.M., O.T., N.S., S.F. and Y.S.; writing—review and editing, B.M., O.T., N.S., S.F. and Y.S.; visualization, B.M., O.T., N.S., S.F. and Y.S.; project administration, N.S. and S.F. All authors have read and agreed to the published version of the manuscript.

## Funding

This research received no external funding.

## Institutional Review Board Statement

Not applicable.

## Informed Consent Statement

Not applicable.

## Data Availability Statement

The data used to support the findings of this study are included within the article. The Credit Card Fraud Detection dataset is available at: https://www.kaggle.com/datasets/mlg-ulb/creditcardfraud (accessed on 15 November 2022); Credit Fraud || Dealing with Imbalanced Datasets is available at: https://www.kaggle.com/code/janiobachmann/credit-fraud-dealing-with-imbalanced-datasets (accessed on 15 November 2022).

## Acknowledgments

The authors would like to thank the Armed Forces of Ukraine for providing security to perform this work. This work was made possible only because of the resilience and courage of the Ukrainian Army.

## Conflicts of Interest

The authors declare no conflict of interest.